\documentclass[11pt,a4paper]{article}
\usepackage{times,latexsym}
\usepackage{url}
\usepackage[T1]{fontenc}

\usepackage[acceptedWithA]{tacl2018v2}

\title{Latent Compositional Representations Improve Systematic Generalization in Grounded Question Answering}

\author{\makecell{Ben Bogin$^{1}$ ~~~~~~~ Sanjay Subramanian$^{2}$   ~~~~~ Matt Gardner$^{2}$~~~~~ Jonathan Berant$^{1,2}$ } \\ 
$^{1}$Tel-Aviv University \hspace{5mm} $^{2}$Allen Institute for AI \\
\texttt{\makecell{\{ben.bogin,joberant\}@cs.tau.ac.il, \{sanjays,mattg\}@allenai.org\\
}}}

\date{}

\usepackage{amsfonts}
\usepackage{amsmath}
\usepackage{svg}
\usepackage{algorithm}
\usepackage[noend]{algpseudocode}
\usepackage{makecell}
\usepackage{booktabs}
\usepackage{rotating}
\usepackage{enumitem}
\usepackage{pifont}
\usepackage[normalem]{ulem}

\newif\ifcomments
\commentstrue
\ifcomments
    \providecommand{\bb}[1]{{\protect\color{olive}{[BB: #1]}}}
    \providecommand{\sanjay}[1]{{\protect\color{magenta}{[SS: #1]}}}
    \providecommand{\jb}[1]{{\protect\color{red}{[JB: #1]}}}
    \providecommand{\matt}[1]{{\protect\color{teal}{[MG: #1]}}}
\else
    \providecommand{\bb}[1]{}
    \providecommand{\sanjay}[1]{}
    \providecommand{\jb}[1]{}
    \providecommand{\matt}[1]{}    
\fi

\newcommand\closure{\textsc{Closure}}
\newcommand\clevr{\textsc{Clevr}}
\newcommand\nobj{n_{\textrm{obj}}}
\newcommand\hdim{h_{\textrm{dim}}}
\newcommand\vc[1]{\textbf{#1}}

\newcommand{\sA}{\mathcal{A}}

\newcommand{\PreserveBackslash}[1]{\let\temp=\\#1\let\\=\temp}
\newcolumntype{C}[1]{>{\PreserveBackslash\centering}p{#1}}
\newcolumntype{R}[1]{>{\PreserveBackslash\raggedleft}p{#1}}
\newcolumntype{L}[1]{>{\PreserveBackslash\raggedright}p{#1}}

\newcommand{\cmark}{\ding{51}}%
\newcommand{\xmark}{\ding{55}}%

\newcommand\correct{{\color{green}\cmark}}
\newcommand\wrong{{\color{red}\xmark}}

\algrenewcomment[1]{\hfill{// #1}}

\begin{document}
\maketitle

\begin{abstract}
Answering questions that involve multi-step reasoning requires decomposing them and using the answers of intermediate steps to reach the final answer. However, state-of-the-art models in grounded question answering often do not explicitly perform decomposition, leading to difficulties in generalization to out-of-distribution examples. In this work, we propose a model that computes a representation and denotation for all question spans in a bottom-up, compositional manner using a CKY-style parser. Our model induces latent trees, driven by end-to-end (the answer) supervision only. We show that this inductive bias towards tree structures dramatically improves systematic generalization to out-of-distribution examples, compared to strong baselines on an arithmetic expressions benchmark as well as on \closure{}, a dataset that focuses on systematic generalization for grounded question answering. On this challenging dataset, our model reaches an accuracy of 96.1\%, significantly higher than prior models that almost perfectly solve the task on a random, in-distribution split.
\end{abstract}

\section{Introduction}

\begin{figure*}
  \centering
  \includegraphics[width=0.75\linewidth]{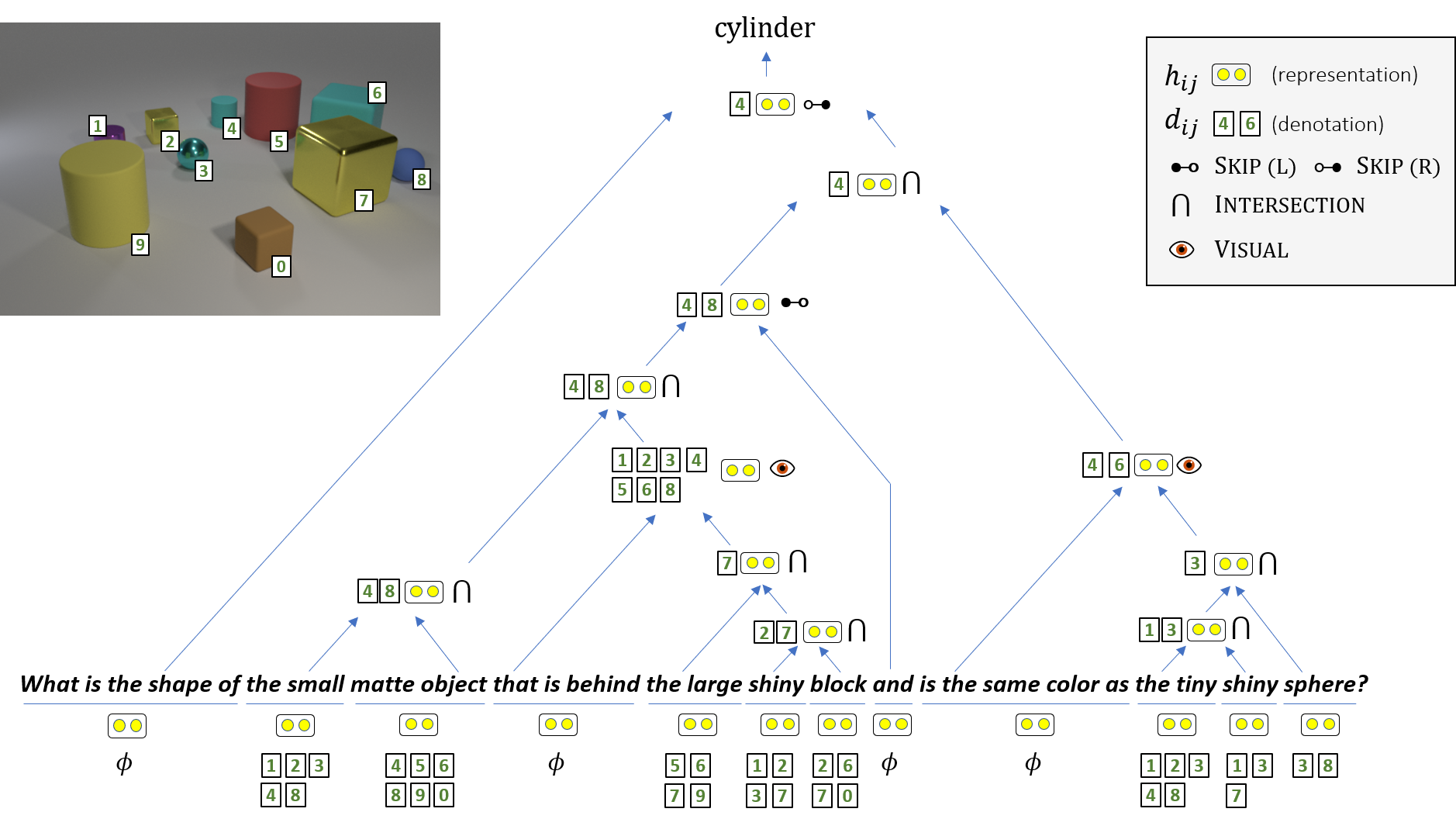}
  \caption{An example from \closure{} illustrating how our model learns a latent structure over the input, where a representation and denotation is computed for every span (for denotation we show the set of objects with probability $>0.5$). For brevity, some phrases were merged to a single node of the tree. For each phrase, we show the split point and module with the highest probability, although all possible split points and module outputs
  are softly computed. 
  \textsc{Skip}(L) and \textsc{Skip}(R) refer to taking the denotation of the left or right sub-span, respectively.
  }
  \label{fig:example}
\end{figure*}

Humans can effortlessly interpret new language utterances, if they are composed of previously-observed primitives and structure \cite{fodor-1988}. Neural networks, conversely, do not exhibit this \emph{systematicity}: while they generalize well to examples sampled from the distribution of the training set, they have been shown to struggle in generalizing to out-of-distribution (OOD) examples that contain new compositions in grounded question answering \cite{bahdanau2018systematic,Bahdanau2019CLOSUREAS} and semantic parsing \cite{finegan-dollak-etal-2018-improving,keysers2020measuring}. Consider the question in Fig.~\ref{fig:example}. This question requires querying object sizes, comparing colors, identifying spatial relations and computing intersections between sets of objects.
Neural models succeed when these concepts are combined in ways that were seen at training time. However, they commonly fail when concepts are combined in new ways at test time.

A possible reason for this phenomenon is the expressivity of architectures such as LSTMs \cite{Hochreiter1997LongSM} and Transformers \cite{vaswani-2017-attention}, where ``global'' representations that depend on the entire input are computed. Contextualizing token representations using the entire utterance potentially lets the model avoid step-by-step reasoning, ``collapse" reasoning steps, and rely on shortcuts \cite{jiang-bansal-2019-avoiding,subramanian-2019-analyzing,Subramanian2020ObtainingFI}. Such failures are revealed when evaluating models for systematic generalization on OOD examples. This stands in contrast to pre-neural log-linear models, where hierarchical representations were explicitly constructed over the input \cite{Zettlemoyer2005LearningTM,liang2013learning}.

In this work, we propose a model for visual question answering (QA) 
that, analogous to these classical pre-neural models, computes for every span in the input question a \emph{representation} and a \emph{denotation}, that is, the set of objects in the image that the span refers to (see Fig.~\ref{fig:example}). 
Denotations for long spans are recursively computed from shorter spans using a bottom-up CKY-style parser \emph{without} access to the entire input, leading to an inductive bias that encourages compositional computation. Because training is done from the final answer only, the model must effectively learn to induce latent trees that describe the compositional structure of the problem. We hypothesize that this explicit grounding of the meaning of sub-spans through hierarchical computation should result in better generalization to new compositions.

We evaluate our approach in two setups: (a) a synthetic arithmetic expressions dataset, and (b) \closure{} \cite{Bahdanau2019CLOSUREAS}, a visual QA dataset focusing on systematic generalization. On a random train/test split (\emph{i.i.d split}), both our model and prior baselines obtain near perfect performance. However, on splits that require generalization to new compositions (\emph{compositional split}) our model dramatically improves performance: on the arithmetic dataset, a Transformer-based model fails to generalize and obtains 2.9\% accuracy, while our model, Grounded Latent Trees (GLT), gets 98.4\%. On \closure{}, our model's accuracy is 96.1\%, 24 absolute points higher than strong baselines and even 19 points higher than models that use gold structures at training time or depend on domain-knowledge.

To conclude, we propose a model with an inherent inductive bias for compositional computation, which leads to gains in systematic generalization, and induces latent structures that are useful for understanding its inner workings. 
This suggests that despite the success of general-purpose architectures built on top of contextualized representations, restricting information flow inside the network can benefit compositional generalization.
Our code and data can be found at \url{https://github.com/benbogin/glt-grounded-latent-trees-qa}.
\section{Compositional Generalization}

Language is mostly compositional; humans can understand and produce a potentially infinite number of novel combinations from a finite set of components \cite{Chomsky1957-CHOSS-2,montague1970aug}. For example, a person would know what a \textit{"winged giraffe"} is even if she's never seen one, assuming she knows the meaning of \textit{``winged''} and \textit{``giraffe''}. This ability, termed \emph{compositional generalization}, is fundamental for building robust models that learn from limited data \cite{Lake2018BuildingMT}.

Neural networks have been shown to generalize well in many language understanding tasks in i.i.d splits \cite{devlin-etal-2019-bert,linzen2020can}.
However, when evaluated on splits that require compositional generalization, a significant drop in performance is observed.
For example, in SCAN \cite{Lake2018GeneralizationWS} and gSCAN \cite{Ruis2020ABF}, synthetically generated commands are mapped into a sequence of actions. When tested on unseen command combinations, models perform poorly. A similar trend was shown in text-to-SQL parsing \citet{finegan-dollak-etal-2018-improving}, where splitting examples by the template of the target SQL query resulted in a dramatic performance drop. SQOOP \cite{bahdanau2018systematic} shows the same phenomena on a synthetic visual task, which tests for generalization over unseen combinations of object properties and relations. This also led to methods that construct compositional splits automatically \cite{keysers2020measuring}.

In this work, we focus on answering complex grounded questions over images. The \clevr{} benchmark \cite{johnson2017clevr} contains synthetic images and questions that require multi-step reasoning, e.g.,  \textit{``Are there any cyan spheres made of the same material as the green sphere?''}. While performance on this task is high, with an accuracy of 97\%-99\% \cite{Perez2018FiLMVR,hudson2018compositional}, recent work \cite{Bahdanau2019CLOSUREAS} introduced \closure{}: a new set of questions with identical vocabulary but different structure than \clevr{}, asked on the same set of images. They evaluated generalization of different models and showed that all fail on a large fraction.

The most common approach for grounded QA 
is based on end-to-end differentiable models such as FiLM \cite{Perez2018FiLMVR}, MAC \cite{hudson2018compositional} LXMERT \cite{tan-bansal-2019-lxmert}, and UNITER \cite{chen2019uniter}.
These models do not explicitly decompose the problem into smaller sub-tasks, and are thus prone to fail on compositional generalization. 
A different approach \cite{yi2018neural,mao2018the} is to parse the image into a symbolic or distributed knowledge graph with objects, attributes and relations, and then parse the question into an executable logical form, which is deterministically executed.
Last, Neural Module Networks (NMNs; \citealt{Andreas2016NeuralMN}) parse the question into an executable program, where execution is learned: each program module is a neural network designed to perform an atomic task, and modules are composed to perform complex reasoning. The latter two model families construct compositional programs and have 
been shown to generalize better on compositional splits 
\cite{bahdanau2018systematic,Bahdanau2019CLOSUREAS} compared to end-to-end models. However, programs are not explicitly tied to question spans, and search over the space of programs is not differentiable, leading to difficulties in training.

In this work, we learn a latent structure for the question and tie each question span to an executable module in a differentiable manner. Our model balances distributed and symbolic approaches: we learn from downstream supervision only and output an inferred tree, describing how the answer was computed.
We base our work on latent tree parsers \cite{le-zuidema-2015-forest,liu-etal-2018-structured,Maillard2019JointlyLS,drozdov-etal-2019-unsupervised} that produce representations for all spans, and softly weight all possible trees. We extend these parsers to answer grounded questions, grounding sub-trees in image objects. 

Closest to our work is \citet{Gupta2018NeuralCD} (GL), who have proposed to compute denotations for each span by constructing a CKY chart. Our model differs in several important aspects: First, while both models compute the denotations of all spans, we additionally compute dense representations for each span using a learned composition function, increasing the expressiveness of our model. Second, we compute the denotations using simpler and more generic composition operators, while in GL the operators are fine-grained and type-driven.
Last, we work with images rather than a knowledge graph, and propose several modifications to the chart construction mechanism to overcome scalability challenges.
\section{Model}
We first give a high-level overview of our proposed Grounded Latent Trees model (GLT) (\S\ref{subsec:overview}), and then explain our model in detail.

\subsection{High-level overview}
\label{subsec:overview}

\paragraph{Problem setup}
Our task is visual QA, where given a question $q = (q_0, \dots, q_{n-1})$ and an image $I$, we aim to output an answer $a \in \sA$ from a fixed set of natural language phrases. 
We train a model from a training set $\{(q^i, I^i, a^i)\}_{i=1}^N$.
We assume we can extract from the image up to $\nobj$ features vectors of objects, and represent them as a matrix $V \in \mathbb{R}^{\nobj \times h_{\textrm{dim}}}$ (details on object detection and representation are in \S\ref{subsec:grounding}).

Our goal is to compute for every question span $q_{ij} = (q_i, \dots, q_{j-1})$ a \emph{representation} $\vc{h}_{ij} \in \mathbb{R}^{\hdim}$ and a \emph{denotation} $\vc{d}_{ij} \in [0,1]^{\nobj}$, interpreted as the probability that the question span refers to each object. 
We compute $\vc{h}_{ij}$ and $\vc{d}_{ij}$ in a bottom-up fashion, using CKY \cite{cocke1969,kasami1965,younger1967}. Algorithm \ref{alg:overview} provides a high-level description of the procedure.
We compute representations and denotations for length-$1$ spans (we use $\vc{h}_{i} = \vc{h}_{i(i+1)}$, $\vc{d}_{i} = \vc{d}_{i(i+1)}$ for brevity) by setting the representation $\vc{h}_{i} = E_{q_i}$ to be the corresponding word representation in an embedding matrix $E$, and grounding each word in the image objects: $\vc{d}_{i} = f_\textrm{ground}(E_{q_i}, V)$
(lines \ref{line:init_start}-\ref{line:init_end}; $f_\textrm{ground}$ is described in \S\ref{subsec:grounding}).
Then, we recursively compute representations and denotations of larger spans (lines \ref{line:rec_start}-\ref{line:rec_end}). Last, we pass the question representation, $\vc{h}_{0n}$, and the weighted sum of the visual representations ($\vc{d}_{0n}V$) through a softmax layer to produce a final answer distribution (line \ref{line:softmax}), using a learned classification matrix $W \in \mathbb{R}^{\sA\times 2h_{\textrm{dim}}}$.

Computing $\vc{h}_{ij}, \vc{d}_{ij}$ for all spans requires overcoming some challenges. Each span representation $\vc{h}_{ij}$ should be a function of two \textit{sub-spans} $\vc{h}_{ik}, \vc{h}_{kj}$. We use the term sub-spans to refer to all adjacent pairs of spans that cover $q_{ij}$: $\{(q_{ik}, q_{kj})\}_{k=i+1}^{j-1}$. However, we have no supervision for the ``correct'' split point $k$. Our model considers all possible split points and learns to induce a latent tree structure from the final answer only (\S\ref{subsec:chart}). We show that this leads to an interpretable compositional structure and denotations that can be inspected at test time.

In \S\ref{subsec:comp_fns} we describe the form of the \emph{composition functions}, which compute
both span representations and denotations from two sub-spans. These functions must be expressive enough to accommodate a wide range of interactions between sub-spans, while avoiding reasoning shortcuts that might hinder compositional generalization.

\begin{algorithm}[t]
{\footnotesize
    \caption{}
    \label{alg:overview}
    \begin{algorithmic}[1]
    \Require question $q$, image $I$, word embedding matrix $E$, visual representations matrix $V$
    \State $H$: tensor holding representations $\vc{h}_{ij}$, $\forall i,j$ s.t. $i < j$
    \State $D$: tensor holding denotations $\vc{d}_{ij}$, $\forall i,j$ s.t. $i < j$
    \For{$i=1 \dots n$} \label{line:init_start}  \Comment{Handle single-token spans}
      \State $\vc{h}_{i} \leftarrow E_{q_i}, \vc{d}_{i} \leftarrow f_\textrm{ground}(E_{q_i}, V)$ (see \S\ref{subsec:grounding}) \label{line:init_end}
    \EndFor
    \For{$l=1 \dots n$} \label{line:rec_start} \Comment{Handle spans of length $l$}
      \State compute $\vc{h}_{ij}$, $\vc{d}_{ij}$ for all entries s.t $j-i = l$ \label{line:rec_end}
    \EndFor
    \State $p(a \mid q, I) \leftarrow \textrm{softmax}(W[\vc{h}_{0n};\vc{d}_{0n}V])$ \label{line:softmax}\\
    \Return $\arg\max_a p(a \mid q, I)$ \label{line:return}
    \end{algorithmic}
    }
\end{algorithm}

\subsection{Grounded chart parsing}
\label{subsec:chart}

We now describe how to compute $\vc{h}_{ij}, \vc{d}_{ij}$ from previously computed representations and denotations. In standard CKY, each constituent over a span $q_{ij}$ is constructed by combining two sub-spans $q_{ik}, q_{kj}$ that meet at a split point $k$. Similarly, we define a representation $\vc{h}_{ij}^k$ conditioned on the split point and constructed from previously-computed representations of two sub-spans:
\begin{equation}
\label{eq:compose_rep}
    \vc{h}_{ij}^k=f_h(\vc{h}_{ik},\vc{h}_{kj}),
\end{equation}
where $f_h(\cdot)$ is a \emph{composition function} (\S\ref{subsec:comp_fns}). 

\begin{figure}
  \centering
  \includegraphics[width=0.85\linewidth]{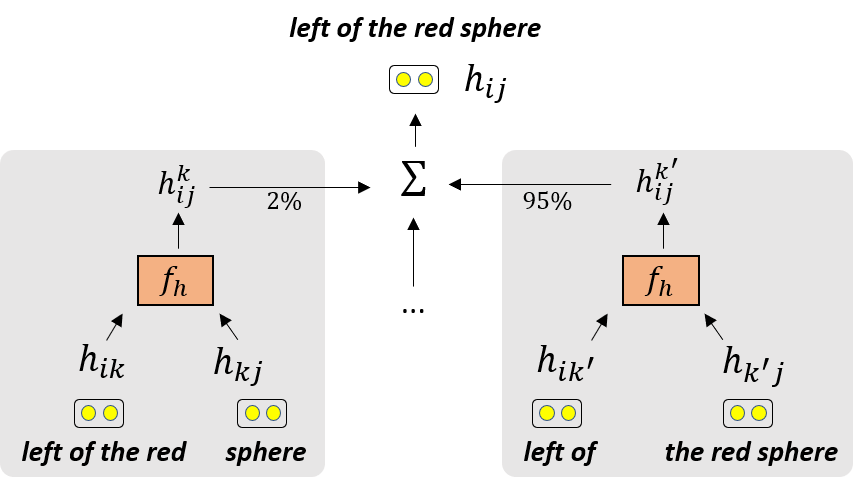}
  \caption{Computing $\vc{h}_{ij}$: we consider all split points and compose pairs of sub-spans using $f_h$. The output is a weighted sum of these representations.}
  \label{fig:h_func}
\end{figure}

Since we want the loss to be differentiable, we do not pick a particular value $k$, but instead use a continuous relaxation. Specifically, we compute $\alpha^k_{ij}$ that represents the probability of $k$ being the split point for the span $q_{ij}$, given the tensor $H$ of computed representations of shorter spans. We then define the representation $\vc{h}_{ij}$ to be the expected representation over all possible split points:
\begin{equation}
\label{eq:hij}
\vc{h}_{ij}=\sum_k {\alpha^k_{ij} \cdot \vc{h}_{ij}^k} = \mathbb{E}_{\alpha^k_{ij}} [\vc{h}_{ij}^k].
\end{equation}
The split point distribution is defined as  $\alpha^k_{ij} \propto \exp(\vc{s}^T \vc{h}_{ij}^k)$, where $\vc{s} \in \mathbb{R}^{\hdim}$ is a parameter vector that determines what split points are likely.
Figure~\ref{fig:h_func} illustrates computing $\vc{h}_{ij}$.

Next, we compute the denotation $\vc{d}_{ij}$ of each span. Conceptually, computing $\vc{d}_{ij}$ can be analogous to $\vc{h}_{ij}$; that is, a function $f_d$ will compute $\vc{d}_{ij}^k$ for every split point $k$, and $\vc{d}_{ij} = \mathbb{E}_{\alpha^k_{ij}}[\vc{d}_{ij}^k]$. However, the function $f_d$ (see \S\ref{subsec:comp_fns}) interacts with the visual representations of all objects and is thus computationally costly. Therefore, we propose a less expressive but more efficient approach, where $f_d(\cdot)$ is applied only once for each span $q_{ij}$. 

Specifically, we compute the expected denotation of the left and right sub-spans of $q_{ij}$:
\begin{align}
\bar{\vc{d}}_{{ij}_L} &= \mathbb{E}_{\alpha^k_{ij}}[\vc{d}_{ik}] && \in \mathbb{R}^{\nobj}, \\
\bar{\vc{d}}_{{ij}_R} &= \mathbb{E}_{\alpha^k_{ij}}[\vc{d}_{kj}] && \in \mathbb{R}^{\nobj}.
\end{align}
If $\alpha^k_{ij}$ puts most probability mass on a single split point $k'$, then the expected denotation will be similar to picking that particular split point.

Now we can compute $\vc{d}_{ij}$ given the expected sub-span denotations and representations with a single application of $f_d(\cdot)$:
\begin{equation}
\label{eq:compose_den}
\vc{d}_{ij} = f_d(\bar{\vc{d}}_{{ij}_L},\bar{\vc{d}}_{{ij}_R},\vc{h}_{{ij}}), 
\end{equation}
which is more efficient than the alternative 
$\mathbb{E}_{\alpha^k_{ij}}[f_d(\vc{d}_{ik},\vc{d}_{kj}, \vc{h}_{ij})]$: $f_d$ is applied $O(n^2)$ times compared to $O(n^3)$ with the simpler solution. This makes training tractable in practice. 

\begin{figure}
  \centering
  \includegraphics[width=0.8\linewidth]{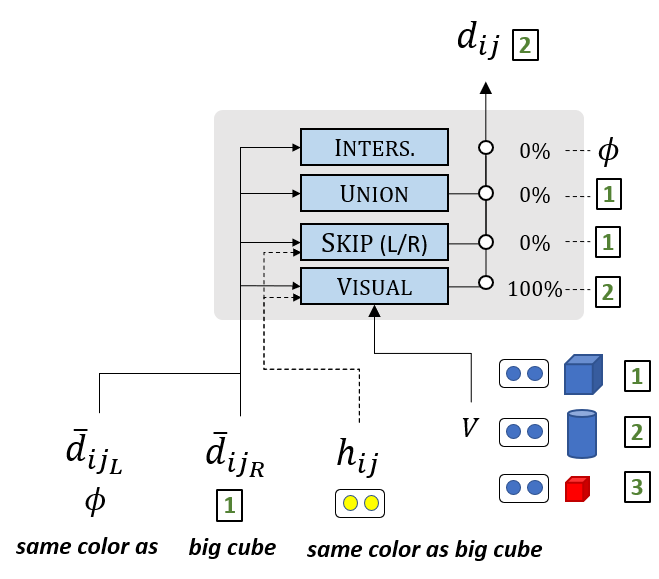}
  \caption{Illustration of how $d_{{ij}}$ is computed. We compute the denotations of all modules, and a weight for each one of the modules. The span denotation is then the weighted sum of the module outputs. }
  \label{fig:d_func}
\end{figure}

\subsection{Composition functions}
\label{subsec:comp_fns}

We now describe the exact form of the composition functions $f_h$ and $f_d$.

\paragraph{Composing representations} We describe the function $f_h(\vc{h}_{ik}, \vc{h}_{kj})$, used to compose the representations of two sub-spans (Eq.~\ref{eq:compose_rep}). The goal of this function is to compose the ``meanings'' of two adjacent spans, \textit{without} access to the rest of the question or the denotations of the sub-spans. For example, composing the representations of \textit{``same''} and \textit{``size''} to a representation for \textit{``same size''}.
At a high-level, composition is based on an attention mechanism.
Specifically, we use attention to form a convex combination of the representations of the two sub-spans (Eq.~\ref{eq:rep_importance}-\ref{eq:rep_weight}), and apply a non-linearity with a residual connection (Eq.~\ref{eq:rep_nonline}).
\begin{align}
    \hspace{-10cm}
    \label{eq:rep_importance}
    &\beta_{ij}^k = \textrm{softmax}\left([\vc{a}_L \vc{h}_{ik}, \vc{a}_R \vc{h}_{kj}]\right) && {\scriptstyle \in \mathbb{R}^2} \\
    \label{eq:rep_weight}
    &\hat{\vc{h}}_{ij}^k = \beta^k_{{ij}_{(1)}} W_L \vc{h}_{ik} +  \beta^k_{{ij}_{(2)}} W_R \vc{h}_{kj} && {\scriptstyle \in \mathbb{R}^{\hdim}} \\
    \label{eq:rep_nonline}
    &f_h(\vc{h}_{ik}, \vc{h}_{kj}) =  \textrm{FF}_{\textrm{rep}}\left(\hat{\vc{h}}_{ij}^k\right) + \hat{\vc{h}}_{ij}^k && {\scriptstyle \in \mathbb{R}^{\hdim}}
\end{align}
where $\vc{a}_L, \vc{a}_R \in \mathbb{R}^{\hdim}$, $W_L, W_R \in \mathbb{R}^{\hdim \times \hdim}$, and $\textrm{FF}_{\textrm{rep}}(\cdot)$ is a linear layer of size ${\hdim} \times {\hdim}$ followed by a non-linear activation.\footnote{We also use Dropout and Layer-Norm \cite{Ba2016LayerN} throughout the paper, omitted for simplicity.}

\paragraph{Composing denotations}
Next, we describe the function $f_d(\bar{\vc{d}}_{{ij}_L},\bar{\vc{d}}_{{ij}_R},\vc{h}_{ij})$, used to compute the span denotation $\vc{d}_{ij}$ (Eq.~\ref{eq:compose_den}). 
This function has access only to words in the span $q_{ij}$ and not to the entire input utterance. We want $f_d(\cdot)$ to support both simple compositions that depend only on the denotations of sub-spans, as well as more complex functions that consider the visual representations of different objects (spatial relations, colors, etc.).

\begin{figure}
  \centering
  \includegraphics[width=0.85\linewidth]{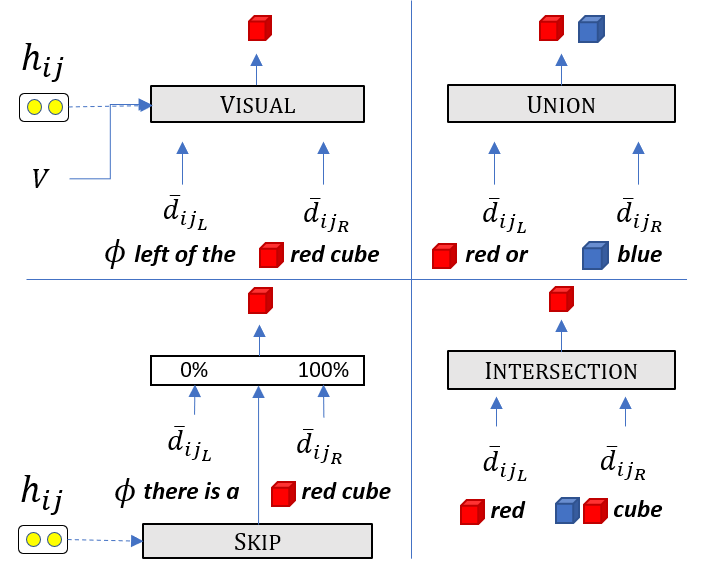}
  \caption{The different modules used with their inputs and expected output.}
  \label{fig:modules}
\end{figure}

We define four modules in $f_d(\cdot)$ for computing denotations and let the model learn when to use each module.
The modules are: \textsc{Skip}, \textsc{Intersection}, \textsc{Union}, and a general-purpose \textsc{Visual} function, where only \textsc{Visual} uses the visual representations $V$. As illustrated in Fig.~\ref{fig:d_func}, each module $m$ outputs a denotation vector $\vc{d}_{{ij}}^m \in [0,1]^{\nobj}$, and the denotation $\vc{d}_{ij}$ is a weighted average of the four modules:
\begin{align}
    &p(m|i,j) \propto \exp(W_\textrm{mod} \vc{h}_{ij})  && \in \mathbb{R}^4 \\
    &\vc{d}_{ij} = \sum_m p(m|i,j)\cdot \vc{d}_{{ij}}^m && \in \mathbb{R}^{\nobj},
\end{align}
where $W_\textrm{mod} \in \mathbb{R}^{\hdim \times 4}$. Next, we define the four modules (see Fig.~\ref{fig:modules}).

\paragraph{\textsc{Skip}} In many cases, only one of the left or right sub-spans have a meaningful denotation: for example, for \textit{``there is a''} and \textit{``red cube''}, we should only keep the denotation of the right sub-span.
To that end, the \textsc{Skip} module weighs the two denotations and sums them ($W_{\textrm{sk}} \in \mathbb{R}^{\hdim \times 2}$):
\begin{align}
    & (c_{{ij}_{(1)}}, c_{{ij}_{(2)}}) = \textrm{softmax}\left(W_{\textrm{sk}} \vc{h}_{ij}\right) && {\scriptstyle \in \mathbb{R}^2} \\
    & \vc{d}_{{ij}}^{\textrm{sk}} = c_{{ij}_{(1)}} \cdot \vc{d}_{{ij}_L} + c_{{ij}_{(2)}} \cdot \vc{d}_{{ij}_R} && {\scriptstyle \in \mathbb{R}^{\nobj}}.
\end{align}

\paragraph{\textsc{Intersection} and \textsc{union}} We define two modules that only use the denotations $\bar{\vc{d}}_{{ij}_L}$ and $\bar{\vc{d}}_{{ij}_R}$. The first corresponds to intersection of two sets, and the second to union:
\begin{align}
    & \vc{d}_{{ij}}^\textrm{int} = \min\left(\bar{\vc{d}}_{{ij}_L}, \bar{\vc{d}}_{{ij}_R}\right) && \in \mathbb{R}^{\nobj} \\
    & \vc{d}_{{ij}}^\textrm{uni} = \max\left(\bar{\vc{d}}_{{ij}_L}, \bar{\vc{d}}_{{ij}_R}\right) && \in \mathbb{R}^{\nobj},
\end{align}
where $\min(\cdot)$ and $\max(\cdot)$ are computed element-wise.

\begin{figure}
  \centering
  \includegraphics[width=0.9\linewidth]{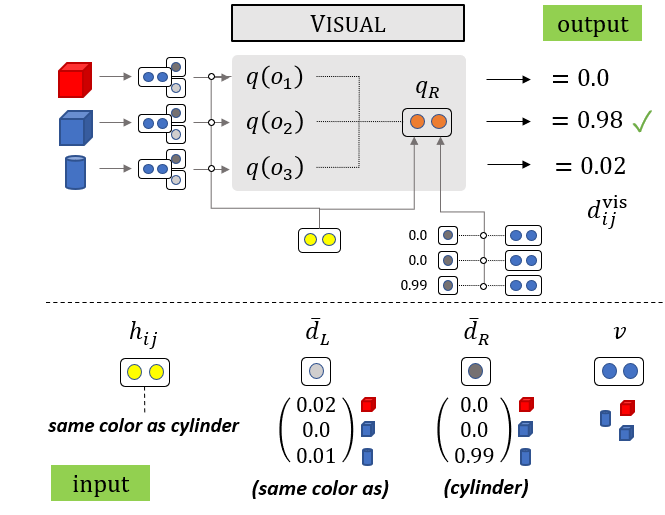}
  \caption{Overview of the \textsc{Visual} module. We compute a  representation for each object, $q(o)$, augmented with the span denotation, representation and visual embeddings, and then compute its dot-product with $q_R$, a representation conditioned on the weighted sum of the right sub-span objects and representations.}
  \label{fig:visual}
\end{figure}

\paragraph{\textsc{Visual}} This module is responsible for visual computations, such as computing spatial relations (\textit{``left of the red sphere''}) and comparing attributes of objects (\textit{``has the same size as the red sphere''}). Unlike other modules, it also uses the visual representations of the objects, $V \in \mathbb{R}^{\nobj \times h_{\textrm{dim}}}$.
For example, for the sub-spans \textit{``left of''} and \textit{``the red object''}, we expect the function to ignore $\bar{\vc{d}}_{{ij}_L}$ (since the denotation of \textit{``left to''} is irrelevant), and return a denotation with high probability for objects that are left to objects with high probability in $\bar{\vc{d}}_{{ij}_R}$.

To determine if an object with index $o$ should have high probability, we need to consider its relation to all other objects. A simple scoring function between object $o$ and another arbitrary object $o_z$ might be $(\vc{h}_{ij} + \vc{v}_{o})^T (\vc{h}_{ij} + \vc{v}_{o_z})$, which will capture the relation between the objects conditioned on the span representation.
However, computing this for all pairs is quadratic in $\nobj$. Instead, we propose a linear alternative that leverages expected denotations of sub-spans.
Specifically, we compute the expected visual representation of the right sub-span and pass it through a feed-forward layer:
\begin{align}
&\overline{v}_R = \bar{\vc{d}}_{R} V \in \mathbb{R}^{\hdim} \label{eq:v_f:2}\\
&\vc{q}_{R} = \textrm{FF}_R \left(W_h\vc{h}_{ij} + \overline{v}_R \right) \in \mathbb{R}^{\hdim} \label{eq:v_f:3}.
\end{align}
We use the right sub-span because the syntax in \clevr{} is mostly right-branching.\footnote{Using the left sub-span instead substantially reduces performance on \clevr{} and \closure{}.} Then, we generate a representation  $\vc{q}(o)$ for every object conditioned on the span representation $\vc{h}_{ij}$, the object probability under the sub-span denotations, and its visual representation. The final object probability is based on the interaction of $\vc{q}(o)$ and $\vc{q}_{R}$:
\begin{align*}
&\vc{q}(o) = \textrm{FF}_\textrm{vis} \left(W_h\vc{h}_{ij} + \vc{v}_o + \bar{\vc{d}}_{L}(o) \vc{s}_L + \bar{\vc{d}}_{R}(o) \vc{s}_R \right) \\
&\vc{d}^{\textrm{vis}}_{ij}(o) = \sigma \left( {\vc{q}(o)}^T \vc{q}_{R} \right),
\end{align*}
where $W_\textrm{h} \in \mathbb{R}^{\hdim \times \hdim}$, $\vc{s}_L, \vc{s}_R \in \mathbb{R}^{\hdim}$ are learned embeddings and $\textrm{FF}_{\mathrm{vis}}$ is a feed-forward layer of size $\hdim \times \hdim$ with a non-linear activation. This is the most expressive module we propose. A visual overview is shown in Fig.~\ref{fig:visual}.

\subsection{Grounding}
\label{subsec:grounding}
In lines~\ref{line:init_start}-\ref{line:init_end} of Algorithm~\ref{alg:overview}, we handle length-$1$ spans. The span representation $\vc{h}_{i}$ is initialized as the corresponding word embedding $E_{q_i}$, and the denotation is computed with a \emph{grounding function}.
A simple implementation for $f_\textrm{ground}$ would use the dot product between the word representation and the visual representations of all objects: $\sigma(\vc{h}_{i}^\top V)$.\footnote{To improve runtime efficiency in the case of a large knowledge-graph, one could introduce a fast filtering function to reduce the number of proposed objects.} But, in the case of co-reference (\emph{``it''}), we should ground the co-referring pronoun in the denotation of a previous span. We now describe this case.

\paragraph{Co-reference}
Sentences such as \textit{``there is a red sphere; what is its material?''} are challenging given a CKY parser, since the denotation of \emph{``its''} depends on a distant span. We propose a simple heuristic
that addresses the case where the referenced object is the denotation of a previous sentence. 
This could be expanded in future work, to a wider array of coreference phenomena.

In every example that comprises two sentences: 
(a) We compute the denotation $\vc{d}^\textrm{first}$ for the entire first sentence as previously described;
(b) We ground each word in the second sentence as proposed above: $\hat{\vc{d}}^\textrm{second}_{i} = \sigma({\vc{h}^\textrm{second}_{i}}^\top V)$; 
(c) For each word in the second sentence, we predict whether it co-refers to $\vc{d}^\textrm{first}$ using a learned gate 
$(r_1, r_2) = \textrm{softmax}\left(\textrm{FF}_{\textrm{coref}}(\vc{h}^\textrm{second}_{i})\right)$, where $\textrm{FF}_{\textrm{coref}} \in \mathbb{R}^{\vc{h}_{\textrm{dim}} \times 2}$. (d) We define $\vc{d}^\textrm{second}_{i} = r_1 \cdot \vc{d}^\textrm{first} + r_2 \cdot \hat{\vc{d}}^\textrm{second}_{i}$.

\paragraph{Visual representations}
Next, we describe how we compute the visual embedding matrix $V$. Two common approaches to obtain visual features are (1) computing a feature map for the entire image and letting the model learn to attend to the correct feature position \cite{hudson2018compositional,Perez2018FiLMVR}; and (2) predicting the locations of objects in the image, and extracting features just for these objects \cite{anderson2018bottom,tan-bansal-2019-lxmert,chen2019uniter}. We use the latter approach, since it simplifies learning over discrete sets, and has better memory efficiency --  the model only attends to a small set of objects rather then the entire image feature map. 

We train an object detector, Faster R-CNN \cite{faster-rcnn}, to predict the location $\textrm{bb}_\textrm{pred} \in \mathbb{R}^{\nobj \times 4}$ of all objects, in the format of bounding boxes (horizontal and vertical positions).

We use gold scene data of 5,000 images from \clevr{} for training (and 1,000 images for validation). In order to extract features for each predicted bounding box, we use the bottom-up top-down attention method of \citet{anderson2018bottom},\footnote{Specifically, we use this version: \url{https://github.com/airsplay/py-bottom-up-attention}.} which produces the features matrix $V_\textrm{pred} \in \mathbb{R}^{\nobj \times D}$, where $D=2048$. Bounding boxes and features are extracted and fixed as a pre-processing step.

Finally, to compute $V$, similar to LXMERT and UNITER we augment object representations in $V_\textrm{pred}$ with their position embeddings, and pass them through a single Transformer self-attention layer to add context about other objects:
$V = \textrm{TransformerLayer}(V_\textrm{pred} W_\textrm{feat} + \textrm{bb}_\textrm{pred} W_{\textrm{pos}})$, where $W_\textrm{feat} \in \mathbb{R}^{D \times \hdim}$ and $W_\textrm{pos} \in \mathbb{R}^{4 \times \hdim}$.

\paragraph{Complexity}
Similar to CKY, we go over all $O(n^2)$ spans in a sentence, and for each span compute $\vc{h}^k_{ij}$ for each of the possible $O(n)$ splits (there is no grammar constant since the grammar has effectively one rule).
To compute denotations $\vc{d}_{ij}$, for all $O(n^2)$ spans, we perform a linear computation over all $\nobj$ objects. Thus, the algorithm runs in time $O(n^3 + n^2 \nobj)$, with similar memory consumption.
This is higher than end-to-end models that do not compute explicit span representations.

\subsection{Training}
\label{subsec:training}
The model is fully differentiable, and we simply maximize the log probability $\log p(a^* \mid q, I)$ of the correct answer $a^*$ (see Algorithm~\ref{alg:overview}).

\section{Experiments}
\label{sec:experiments}
In this section, we evaluate our model on both in-distribution and out-of-distribution splits.

\subsection{Arithmetic expressions}

\begin{figure}
  \centering
  \includegraphics[width=0.7\linewidth]{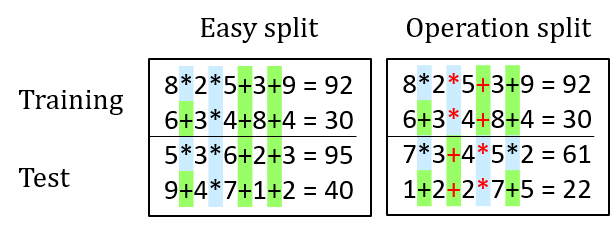}
  \caption{Arithmetic expressions: unlike the easy setup, we evaluate models on expressions with operations ordered in ways unobserved at training time. Flipped operator positions are in red.
  }
  \label{fig:math}
\end{figure}

It has been shown that neural networks can be trained to perform numerical reasoning \cite{Zaremba2014LearningTE,Kaiser2016neuralgpu,trask2018neural,Geva2020InjectingNR}. However, models are often evaluated on expressions that are similar to the ones they were trained on, where only the numbers change. To test generalization, we create a simple dataset and evaluate on two splits that require learning the correct operator precedence. In the first split, sequences of operators that appear at test time do not appear at training time. In the second split, the test set contains longer sequences compared to the training set.

We define an arithmetic expression as a sequence containing $n$ numbers with $n-1$ arithmetic operators between each pair. The answer $a$ is the result of evaluating the expression. 

\paragraph{Evaluation setups} The sampled operators are addition and multiplication, and we use only expressions where $a \in \{0, 1, \dots, 100\}$ to train as a multi-class problem. At training, we randomly pick the length $n$ to be up to $n_{\textrm{train}}$, and at test time we choose a fixed length $n_{\textrm{test}}$. We evaluate on three setups, (a) \emph{Easy split}: we choose $n_{\textrm{train}}=n_{\textrm{test}}=8$, and randomly sample operators from a uniform distribution for both training and test examples. In this setup, we only check that the exact same expression is not shared between the training and test set. (b)  \emph{Operation split}: we randomly pick 3 positions, and for each one randomly assign exactly one operator that will appear at training time. On the test set, the operators in all three positions are flipped, so that they contain the unseen operator (Fig.~\ref{fig:math}). The same lengths are used as in the easy split. (c) \emph{Length split}: we train with $n_{\textrm{train}}=8$ and test with $n_{\textrm{test}}=10$. Examples for all setups are generated on-the-fly for 3 million steps (batch size is $100$).

\paragraph{Models} We compare GLT to a standard Transformer, where the input is the expression, and the output is predicted using a classification layer over the \texttt{[CLS]} token. All models are trained with cross-entropy loss given the correct answer.

For both models, we use an in-distribution validation set for hyper-parameter tuning. 
Since here we do not have an image, we only compute $\vc{h}_{ij}$ for all spans, and define $p(a \mid q)=\textrm{softmax}(W\vc{h}_{0n})$.

GLT layers are almost entirely recurrent, that is, the same parameters are used to compute representations for spans of all lengths. The only exception are layer-normalization parameters, which are not shared across layers. Thus, at test time when processing an expression longer than observed at training time, we use the layer-normalization parameters (total of $2\cdot \hdim$ parameters per layer) from the longest span seen at training time.\footnote{Removing layer normalization leads to improved accuracy of 99\% on the arithmetic expressions length split, but training convergence on \clevr{} becomes too slow.}

\paragraph{Results} Results are reported in Table~\ref{tab:math-results}. We see that both models almost completely solve the in-distribution setup, but on out-of-distribution splits the Transformer performs poorly, while GLT shows only a small drop in accuracy. 

\begin{table}[]
\scriptsize
\centering
\begin{tabular}{llll}
\toprule
     & Easy split & Op. split & Len. split \\
\midrule
Transformer & $\textbf{100.0} \pm 0.0$         & $2.9 \pm 1.1$         & $10.4 \pm 2.4$ \\  
GLT  & $99.9 \pm 0.2$         & $\textbf{98.4} \pm 0.7$    & $\textbf{94.9} \pm 1.1$    \\           
\bottomrule
\end{tabular}
\caption{Arithmetic expressions results for easy split, operation split, and length split.
}
\label{tab:math-results}
\end{table}

\begin{table}[]
\centering
\scriptsize
\begin{tabular}{lll}
\toprule
& \clevr{} & \closure{} \\ \midrule
MAC                  & 98.5    & 72.4   \\
FiLM                 & 97.0     & 60.1       \\
\textbf{GLT (our model)} & 99.1   & \textbf{96.1} $\pm$ 2.5    \\
\midrule
NS-VQA $\dagger$ $\mp$ & \textbf{100}     & 77.2   \\ \midrule
PG+EE (18K prog.) $\dagger$ & 95.4     & -   \\
PG-Vector-NMN $\dagger$ & 98.0     & 71.3     \\  \midrule
GT-Vector-NMN $\dagger$ $\ddagger$ & 98.0     & 94.4      \\
\bottomrule
\end{tabular}
\caption{Test results for all models on \clevr{} and \closure{}. $\dagger$ stands for models trained with gold programs, $\ddagger$ for oracle models evaluated using gold programs, and $\mp$ for deterministically executed models that depend on domain-knowledge for execution (execution is not learned).}
\label{tab:clevr-results}
\end{table}

\subsection{\clevr{} and \closure{}}
\label{subsec:clevr}
We evaluate performance on grounded QA using \clevr{} \cite{johnson2017clevr}, consisting of 100,000 synthetic images with multiple objects of different shapes, colors, materials and sizes. 864,968 questions were generated using 80 different templates, including simple questions (\textit{``what is the size of red cube?''}) and questions requiring multi-step reasoning (Figure~\ref{fig:example}). 
The split in this dataset is i.i.d: the same templates are used for the training, validation, and test sets.

To test compositional generalization when training on \clevr{}, we use the \closure{} dataset \cite{Bahdanau2019CLOSUREAS}, which includes seven new question templates, with a total of 25,200 questions, asked on the \clevr{} validation set images. The templates are created by taking referring expressions of various types from \clevr{} and combining them in novel ways. 

A problem found in \closure{} is that sentences from the template \texttt{embed\_mat\_spa} are ambiguous. For example, in the question \textit{``Is there a sphere on the left side of the cyan object that is the same size as purple cube?''}, the phrase \textit{``that is the same size as purple cube''} can modify either \emph{``the sphere''} or \emph{``the cyan object''}, but the answer in \closure{} is always the latter. Therefore, we deterministically compute both of the two possible answers and keep two sets of question-answer pairs of this template for the entire dataset. We evaluate models\footnote{We update the scores on \closure{} for MAC, FiLM and GLT due to this change in evaluation. The scores for the rest of the models were not affected.} on this template by taking the maximum score over these two sets (such that models must be consistent and choose a single interpretation for the template to get a perfect score).

\begin{table}[]
\centering
\scriptsize
\begingroup
\setlength{\tabcolsep}{5.5pt} 
\begin{tabular}{llllll}
\toprule
& \clevr{} & \closure{} & C.Humans \\ \midrule
\textbf{GLT}  & 99.1 $\pm$ 0.0 & 96.1 $\pm$ 2.7 & 75.8 $\pm$ 0.7   \\
\midrule
\{\textsc{Visual}, \textsc{Skip}\} & 98.9 $\pm$ 0.1     & 91.8 $\pm$ 10.1      & 76.3 $\pm$ 0.6\\
\{\textsc{Visual}\} & 99.0 $\pm$ 0.1      & 83.8 $\pm$ 1.8   & 72.0 $\pm$ 1.6\\
with stop-words & 99.3     & 93.2    & 75.4  \\
contextualized input & 98.9 $\pm$ 0.1     & 87.5 $\pm$ 12.1  & 76.1 $\pm$ 0.9\\
non-compositional & 97.7 $\pm$ 2.0     & 83.5 $\pm$ 0.5 & 75.8 $\pm$ 2.3 \\
\bottomrule
\end{tabular}
\endgroup
\caption{Ablations on the development sets of \clevr{}, \closure{} and \clevr{}-Humans.}
\label{tab:clevr-ablations}
\end{table}

\paragraph{Baselines} We evaluate against the baselines from \citet{Bahdanau2019CLOSUREAS}. The most comparable baselines are MAC \cite{hudson2018compositional} and FiLM \cite{Perez2018FiLMVR}, which are differentiable and do not use program annotations. 
We also compare to NMNs that require a few hundred program examples for training. We show results for PG+EE \cite{johnson2017inferring} and an improved version, PG-Vector-NMN \cite{Bahdanau2019CLOSUREAS}.
Last, we compare to NS-VQA, which in addition to parsing the question, also parses the scene into a knowledge graph. 
NS-VQA requires additional gold data to parse the image into a knowledge graph based on data from \clevr{} (colors, shapes, locations, etc.).

\paragraph{Setup}
Baseline results are taken from previous papers \cite{Bahdanau2019CLOSUREAS,hudson2018compositional,yi2018neural,johnson2017inferring}, except for MAC and FiLM on \closure{}, which we re-executed due to the aforementioned evaluation change.
For GLT, we use \clevr{}'s validation set for hyper-parameter tuning and early-stopping, for a maximum of 40 epochs. We run 4 experiments to compute mean and standard deviation on \closure{} test set.

Because of our model's run-time and memory demands (see \S\ref{subsec:grounding}), we found that running on \clevr{} and \closure{}, where question length goes up to 43 tokens, can be slow. Thus, we delete function words that typically have empty denotations and can be safely skipped,\footnote{The removed tokens are punctuations, `the', `there', `is', `a', `as', `of', `are', `other', `on', `that'.} reducing the maximum length to 25. We run a single experiment where stop words were not removed and report results in Table~\ref{tab:clevr-ablations} (with stop-words), showing that performance only mildly change.

\paragraph{\clevr{} and \closure{}}
In this experiment we compare results on i.i.d and compositional splits. Results are in Table~\ref{tab:clevr-results}. We see that GLT performs well on \clevr{} and gets the highest score on \closure{}, improving by almost 24 points over comparable models, and by 19 points over NS-VQA, outperforming even the oracle GT-Vector-NMN which uses gold programs at \textbf{test time}.

\paragraph{Removing modules} We defined (\S\ref{subsec:comp_fns}) two modules specifically for \clevr{}, \textsc{Intersection} and \textsc{union}. We evaluate performance without them, keeping only \textsc{Visual} and \textsc{Skip}, and show results in Table~\ref{tab:clevr-ablations}, observing only a moderate loss in accuracy and generalization. Removing these modules leads to more cases where the \textsc{Visual} function is used, effectively performing intersection and union as well. While the drop in accuracy is small, this model is harder to interpret since \textsc{Visual} now performs multiple functions.

Next, we evaluate performance when training only with the  \textsc{Visual} module. Results show that even a single high-capacity module is enough for in-distribution performance, but generalization to \textsc{Closure} now substantially drops to 83.8.

\begin{table}[]
\centering
\scriptsize
\begin{tabular}{lll}
\toprule
 & \closure{} FS & C.Humans \\ \midrule
MAC                  & 90.2 & \textbf{81.5}   \\
FiLM                 &  -    & 75.9       \\ 
\textbf{GLT (our model)} & \textbf{97.4} $\pm$ 0.3 & 76.2 \\
\midrule
NS-VQA               & 92.9 & 67.0 \\ \midrule
PG-Vector-NMN  & 88.0 & - \\
PG+EE (18K prog.) & -  & 66.6   \\
\bottomrule
\end{tabular}
\caption{Test results in the few-shot setup and for \clevr{}-Humans.}
\label{tab:closure-fs-human}
\end{table}

\paragraph{Contextualizing question tokens} 
We hypothesized that the fact that question spans do not observe the entire input might aid composition generalization. To check this, we pass the question tokens $q$ through a Transformer self-attention layer, to get contextualized tokens embeddings $E_{q_i}$. We show in Table~\ref{tab:clevr-ablations} that performance on \clevr{} remains the same, but on \closure{} it drops to 87.5.

\paragraph{Non-compositional representations} To assess the importance of \emph{compositional} representations, we replace $\vc{h}_{ij}$ (Eq.~\ref{eq:hij}) with a sequential encoder. Concretely, we set the representation of the span $q_{ij}$ to be $\vc{h}_{ij} = \textrm{BiLSTM}(q_{ij})$, the output of a single layer bi-directional LSTM which is given the tokens of the span as input. We see a drop in both \clevr{} and \closure{} (experiments were more prone to overfitting), showing the importance of compositional representations.

\paragraph{Few-shot} In the few-shot (FS) setup, we train with additional 252 out-of-distribution examples: 36 questions for each \closure{} template. Similar to \citet{Bahdanau2019CLOSUREAS}, we take a model that was trained on \clevr{} and fine-tune adding oversampled \closure{} examples (300x) to the original training set. To make results comparable to \citet{Bahdanau2019CLOSUREAS}, we perform model selection based on \closure{} validation set, and evaluate on the test set. As we see in Table~\ref{tab:closure-fs-human}, GLT gets the best accuracy. If we perform model selection based on \clevr{} (the preferred way to evaluate in the OOD setup, \citealt{Teney2020OnTV}),
accuracy remains similar at  97.0 $\pm$ 2.0 (and is still highest).

\begin{figure}
  \centering
  \includegraphics[width=0.95\linewidth]{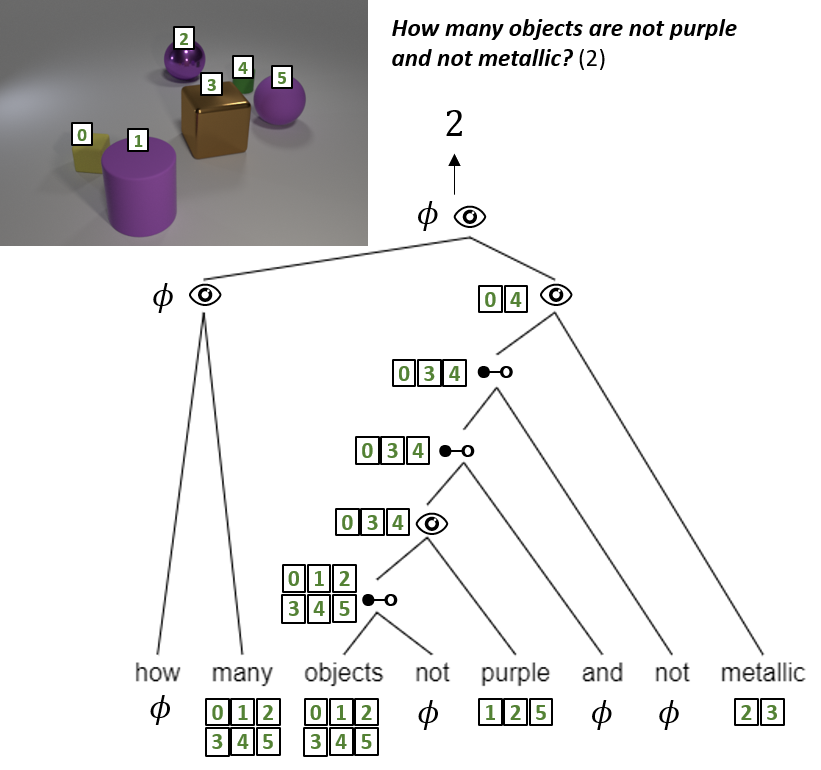}
  \caption{An example from \clevr{}-Humans. The model learned to negate (\textit{``not''}) using the \textsc{Visual} module (negation is not part of \clevr{}).
  }
  \label{fig:ex-negation}
\end{figure}

\paragraph{CLEVR-Humans} To test performance on real language, we use \clevr{}-Humans \cite{johnson2017inferring}, which includes 32,164 questions over images from \clevr{}. These questions, asked by humans, contain words and reasoning steps that were unseen in \clevr{}. We train on \clevr{} and fine-tune on CLEVR-Human, similar to prior work. We use GloVe embeddings \cite{pennington-etal-2014-glove} for words unseen in \clevr{}. Table~\ref{tab:closure-fs-human} shows that GLT gets better results than models that use programs, showing its flexibility to learn new concepts and phrasings. It is also comparable to \textsc{FiLM}, but gets lower results than \textsc{MAC} (see error analysis below).

\subsection{Error analysis} We sampled 25 errors from each dataset (\clevr{}, \closure{}, and \clevr{}-Humans) for analysis. On \textbf{\clevr{}}, most errors (84\%) are due to problems in \emph{visual processing} such as  grounding 
the word \textit{``rubber''} to a metal object, problems in bounding box prediction or questions that require subtle spatial reasoning, such as identifying if an object is left to another object of different size, when they are at an almost identical x-position. The remaining errors (16\%) are due to failed comparisons of numbers or attributes (\textit{``does the red object have the same material as the cube''}). 

On \textbf{\closure{}}, 60\% of the errors were similar to \clevr{}, e.g. problematic visual processing or failed comparisons. In 4\% of cases, the execution of the \textsc{Visual} module was wrong, e.g., it collapsed two reasoning steps (both intersection and finding objects of same shape), but did not output the correct denotation.  Other errors (36\%) are in the predicted latent tree, where the model was uncertain about the split point and softly predicted more than one tree, resulting in a wrong answer. In some cases (16\%) this was due to question ambiguity (see \S\ref{subsec:clevr}), and in others cases the cause was unclear (e.g., for the phrase \textit{``same color as the cube''} the model gave a similar probability for the split after \textit{``same''} and after \textit{``color''}, leading to a wrong denotation of that span).

\begin{table*}[!t]
\centering
\scriptsize
\begin{tabular}{L{3.5cm}p{9.5cm}L{1.6cm}}
\toprule
{\bf Category} & {\bf Examples} & {\bf Correct}\\
 \midrule
  Negation & 
    -- How many objects are not shiny? \correct \newline
    -- What color is the cylinder that is not the same \uwave{colr} as the sphere? \wrong
 & 1/2 (50\%) \\
 \midrule
  Spelling mistakes & 
    -- Are there two green \uwave{cumes?} \correct \newline
    -- How many rubber \uwave{spehres} are there here? \wrong
 & 1/5 (20\%) \\
 \midrule
  Superlatives & 
    -- What color is the object furthest to the right? \correct \newline
    -- What shape is the smallest matte object? \correct
 & 8/10 (80\%) \\
 \midrule
  Visual Concepts: obscuring, between & 
    -- Is the sphere the same color as the object that is obscuring it? \correct \newline
    -- What color is the object in between the two large cubes? \correct
 & 5/7 (71\%) \\
 \midrule
  Visual Concepts: reflection, shadow & 
    -- Which shape shows the largest reflection \wrong \newline
    -- Are all of the objects casting a shadow? \wrong
 & 0/2 (0\%) \\
 \midrule
  Visual Concepts: relations & 
    -- What color is the small ball near the brown cube? \correct \newline
    -- What is behind and right of the cyan cylinder? \wrong
 & 3/8 (38\%) \\
 \midrule
  All quantifier & 
    -- Are all the spheres the same size? \correct \newline
    -- Are all the cylinders brown? \wrong
 & 10/12 (83\%) \\
 \midrule
  Counting abstract concepts & 
    -- How many different shapes are there? \correct \newline
    -- How many differently colored cubes are there? \wrong
 & 1/4 (25\%) \\
 \midrule
  Complex logic & 
    -- if these objects were lined up biggest to smallest, what would be in the middle? \correct \newline
    -- if most of the items shown are shiny and most of the items shown are blue, would it be fair to say most of the items are shiny and blue? \wrong
 & 3/4 (75\%) \\
 \midrule
  Different question structure & 
    -- Are more objects metallic or matte? \correct \newline
    -- Each shape is present 3 times except for the \wrong
 & 1/2 (50\%) \\
 \midrule
  Uniqueness & 
    -- What color object is a different material from the rest? \correct \newline
    -- What color is the object that does not match the others? \wrong
 & 1/2 (50\%) \\
 \midrule
  Long-tail concepts & 
    -- Can you roll all the purple objects? \correct \newline
    -- How many of these things could be stacked on top of each other? \wrong
 & 2/5 (40\%) \\
 \midrule
  Operators used differently than \clevr{} & 
    -- Are the large cylinders the same color? \correct \newline
    -- Are there more rubber objects than matte cylinders and green cubes? \wrong
 & 2/5 (40\%) \\
 \midrule
  Same concepts as \clevr{}, possibly different phrasing & 
    -- What color is the cube directly in front of the blue cylinder? \correct \newline
    -- What color is the little cylinder? \wrong
 & 71/80 (89\%) \\
 \bottomrule
\end{tabular}
\caption{Categorization of 150 \clevr{}-Humans examples, together with the accuracy of the GLT model (for many categories, the sample is too small to make generalizations on model performance). Spelling mistakes are shown with a \uwave{wavy underline}. Some examples were labeled with multiple categories. Six examples with gold-label mistakes were ignored.}
\label{tab:clevr-human-analysis}
\end{table*}

On \textbf{CLEVR-Humans}, the model successfully learned certain new ``concepts'' such as colors (\textit{``gold''}), superlatives (\textit{``smallest'', ``furthest''}), relations (\textit{``next to''}, \textit{``between''}), negation (see Fig.~\ref{fig:ex-negation}) and the \textit{``all``} quantifier. It also answered correctly questions that had different style from \clevr{} (\textit{``Are there more blocks or balls?''}). However, the model is not always robust to these new concepts, and fails in other cases that require counting of abstract concepts (such as colors or shapes) or that use \clevr{}'s operators differently than the original uses. See Table~\ref{tab:clevr-human-analysis} for our manually performed analysis and examples over 150 \clevr{}-Humans validation questions.

\subsection{Limitations}
While some of the errors shown in Table~\ref{tab:clevr-human-analysis} can be improved with careful design of modules architectures, improvement of the image features, or by introducing a pre-training mechanism, some question structures are inherently more difficult to correctly answer using our model. We describe two main limitations.

One key issue is discrepancy between the text of the question and  the actual reasoning steps that are required to correctly answer it. For example, in the question \textit{``What is the most common shape?''}, 
the phrase \emph{``most common''} entails grouping and counting shapes, and then taking an \texttt{argmax} as an answer. This complex reasoning cannot be constructed compositionally from question spans.

The second issue is the supported types of phrase denotations. GLT only grounds \emph{objects} in the image to a phrase, however, in some cases denotations should be a number or an attribute. For example, in comparison questions (\textit{``is the number of cubes higher than the number of spheres?''}) a model would ideally have a numerical denotation for each group of objects.

Importantly, these limitations do not prevent the model from answering questions correctly, since even when denotations do not accurately represent reasoning, the model can still ``fall back'' to the flexible answer function, which can take the question and image representations and, in theory, perform anyf needed computation. In such cases, the model will not be interpretable, similar to other high-capacity models such as Transformers. In future work, we will explore combining the compositional generalization abilities of GLT with the advantages of high-capacity architectures.

\subsection{Interpretability}
A key advantage of latent trees is interpretability --  one can analyze the structure of a given question. Next, we quantitatively inspect the quality of the intermediate outputs produced by the model by (a) comparing the intermediate outputs to ground truth and (b) assessing how helpful these outputs are for humans.

\paragraph{Evaluating intermediate outputs} We assess how accurate our intermediate outputs are compared to ground truth trees. Evaluating against ground truth trees is not trivial, since for a given question, there could be many ``correct'' trees  -- that is, trees that represent a valid way to answer the question. For example, the two possible ways to split the phrase \emph{``large shiny block''} are both potentially valid. Thus, even if we had one ground-truth tree for each question, comparing predicted trees to ground-truth trees might not be reliable.

Instead, we exploit the fact that for the synthetic questions in \clevr{}, it is possible to define a set of certain constituents that \emph{must} appear in any valid tree. For example, phrases starting with \emph{``the''}, followed by adjectives and then a noun (\emph{``the shiny tiny sphere''}), should always be considered a constituent. We use such manually defined rules to extract the set $c_q$ of obligatory constituents for any question $q$ in \clevr{} and \closure{} validation sets.

\begin{table}[]
\centering
\scriptsize
\begin{tabular}{lll}
\toprule
& \clevr{} & \closure{} \\ \midrule
Constituents (\%)             & 83.1    & 81.6   \\
Denotations ($F_1$)              & 95.9    & 94.7       \\
\bottomrule
\end{tabular}
\caption{Evaluation of the intermediate outputs, showing both the recall of expected constituents (\%) and average $F_1$ of predicted objects.
}
\label{tab:ground-truth-trees}
\end{table}

We compute a recall score for each question $q$, by producing predicted trees (we greedily take the maximum probability split in each node), extracting a set $\hat{c}_q$ of all constituents, and then computing 
the proportion of constituents in $c_q$ that are also in $\hat{c}_q$. We show results in Table~\ref{tab:ground-truth-trees} for both \clevr{} and \closure{}. We see that 81.6-83.1\% of the expected constituents are correctly output by GLT, for \clevr{} and \closure{} respectively. Next, we sample and analyze 25 cases from each dataset where the expected constituents were not output. We observe that almost all missed constituents actually \textit{were} output as a constituent by the model, but with an additional prefix that caused a wrong split. For example, instead of the expected constituent \textit{``tiny green objects``} we found the constituent \textit{``any tiny green objects''} (split after \textit{``tiny''}). These mistakes did not cause any
error in
module execution, except for one case in \closure{}.

Additionally, we obtain the gold set of objects for each constituent in $c_q$ by deterministically mapping each constituent to a program that we execute on the scene. For each constituent in $c_q \cap \hat{c}_q$, we compare the predicted set of objects (objects with probability above 50\%) with the gold set of objects, and report the average $F_1$. As can be seen in Table~\ref{tab:ground-truth-trees}, reported score for \clevr{} is 95.9\% and for \closure{} it is 94.7\%.

\paragraph{Human Interpretability} Next, we want to assess how useful are the intermediate outputs produced by the model for humans. We perform the ``forward prediction'' test \cite{hu2018explainable} that evaluates interpretability by showing humans questions accompanied with intermediate outputs, and asking them to predict if the model will be correct. This test is based on the assumption that  humans can more easily predict the accuracy of an interpretable model versus a non-interpretable one. Thus, we also ask a separate group of participants to predict the accuracy \emph{without} showing them the intermediate outputs for comparison.

\begin{table}[]
\centering
\scriptsize
\begin{tabular}{llll}
\toprule
& \clevr{} & \closure{} & C.Humans \\ \midrule
A - w/o tree          & 61.4\%    & 58.1\% & 52.3\%          \\
B - w/ tree            & 62.5\%    & 82.5\% & 77.5\%   \\
\bottomrule
\end{tabular}
\caption{Results for the ``forward prediction'' test. Group A sees only the question, image and gold answer, group B additionally sees the denotation tree.}
\label{tab:forward-predict}
\end{table}

We show each of the 20 participants 12 questions evenly split into \clevr{}, \closure{} and \clevr{}-Humans, where for each dataset we show two correct and two incorrect examples per person. We assign 11 participants to the baseline group (group A) that sees only the question, image, and gold answer, and 9 participants to the tested group (group B) which are also given the predicted denotation tree of our model, similar to Figure~\ref{fig:example}. Both groups are given the same 8 ``training'' examples of questions, with the predicted answer of the model along with the gold answer, to improve their understanding of the task. In these training examples, group B participants also see example denotation trees, along with a basic explanation of how to read those.

We show results in Table~\ref{tab:forward-predict}. We see that for \closure{} and \clevr{}-Humans, accuracy for group B is significantly higher ($p<0.05$), but for \clevr{} results are similar. We observe that in \clevr{}, where accuracy is 99.1\%, wrong predictions are mostly due to errors in the less interpretable \textsc{Visual} module, while for \closure{} and \clevr{}-Humans errors are more often due to wrong selection of constituents and modules, which can be spotted by humans.
\section{Conclusion}
Developing models that generalize to compositions that are unobserved at training time has recently sparked substantial research interest. In this work we propose a model with a strong inductive bias towards compositional computation, which leads to large gains in systematic generalization and 
provides an interpretable structure that can be inspected by humans. Moreover, Our model also obtains high performance on real language questions (\clevr{}-humans). In future work, we will investigate the structures revealed by our model in other grounded QA setups, and will allow the model freedom to incorporate non-compositional signals, which go hand in hand with compositional computation in natural language.

\section*{Acknowledgements}
This research was partially supported by  The Yandex Initiative for Machine Learning, and the European Research Council (ERC) under the European Union Horizons 2020 research and innovation programme (grant ERC DELPHI 802800). We thank Jonathan Herzig and the anonymous reviewers for their useful comments. This work was completed in partial fulfillment for the Ph.D degree of Ben Bogin.

\bibliography{tacl2018}
\bibliographystyle{acl_natbib}

\clearpage

\appendix
\section{Output Examples}
\label{app:examples}
We show 4 additional examples of our model outputs, along with the induced trees and denotations in the following pages.

\begin{sidewaysfigure*}
  \centering
  \includegraphics[width=\linewidth]{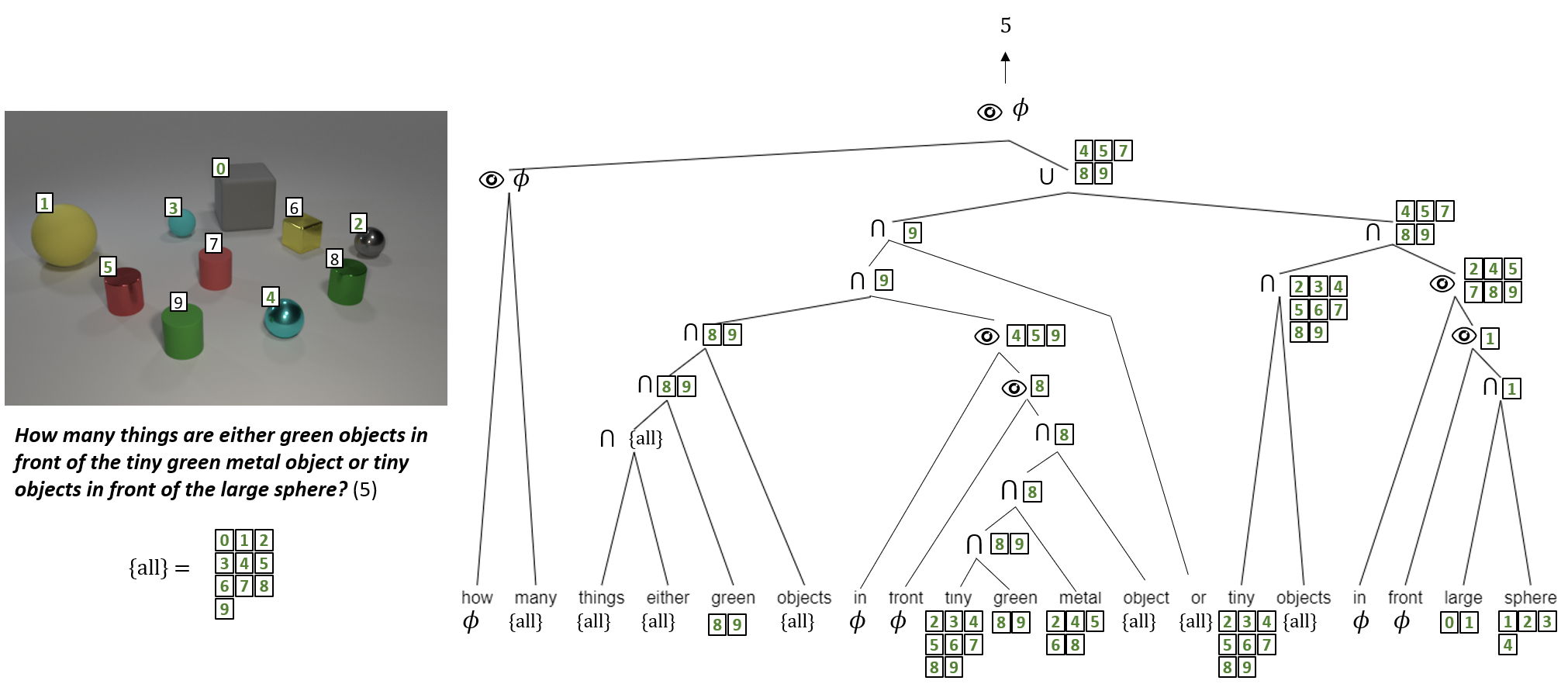}
  \caption{An example from \closure{}. $\cup$ stands for the \textsc{Union} module.
  }
  \label{fig:ex3}
\end{sidewaysfigure*}

\begin{figure*}
  \centering
  \includegraphics[width=\linewidth]{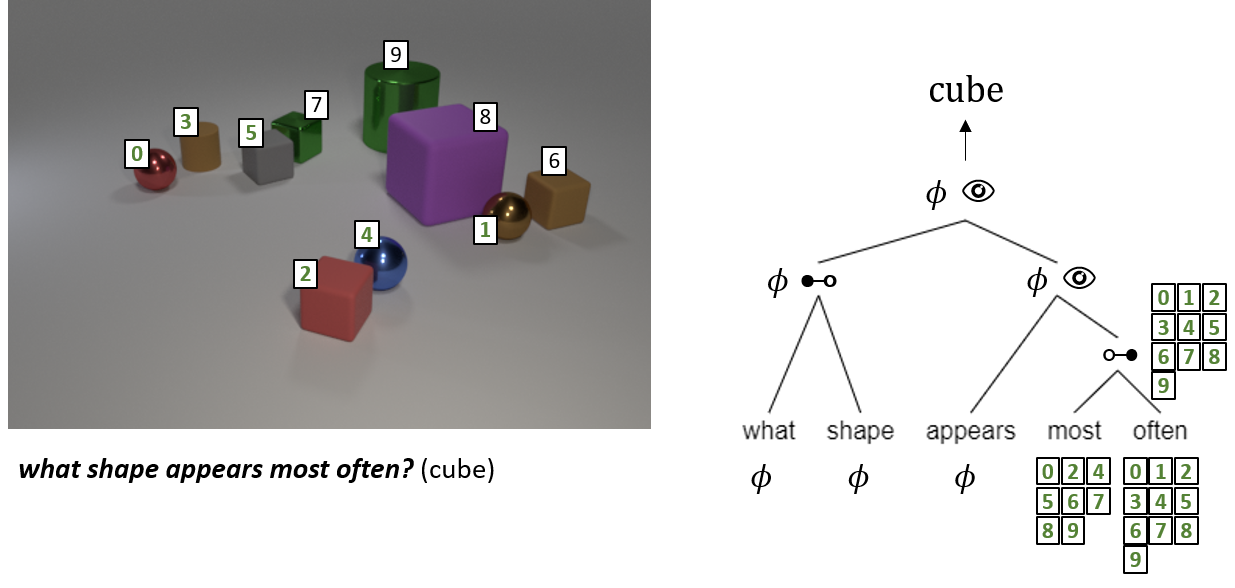}
  \caption{An example from \clevr{}-Humans. This question requires reasoning steps that are not explicitly mentioned in the input. This results in a correct answer but non-interpretable output.
  }
  \label{fig:ex-hard}
\end{figure*}

\begin{figure*}
  \centering
  \includegraphics[width=\linewidth]{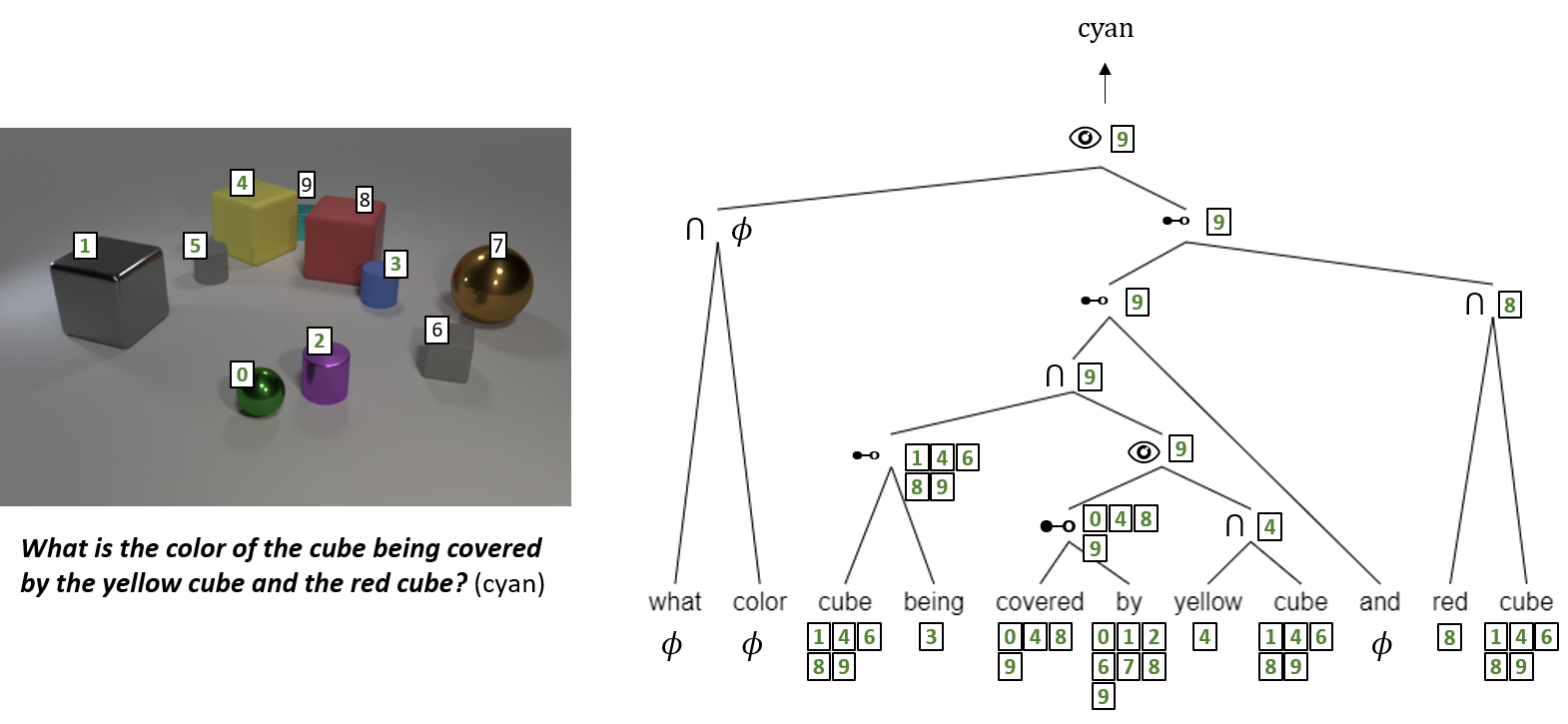}
  \caption{An example from \clevr{}-Humans.
  }
  \label{fig:ex4}
\end{figure*}

\begin{figure*}
  \centering
  \includegraphics[width=\linewidth]{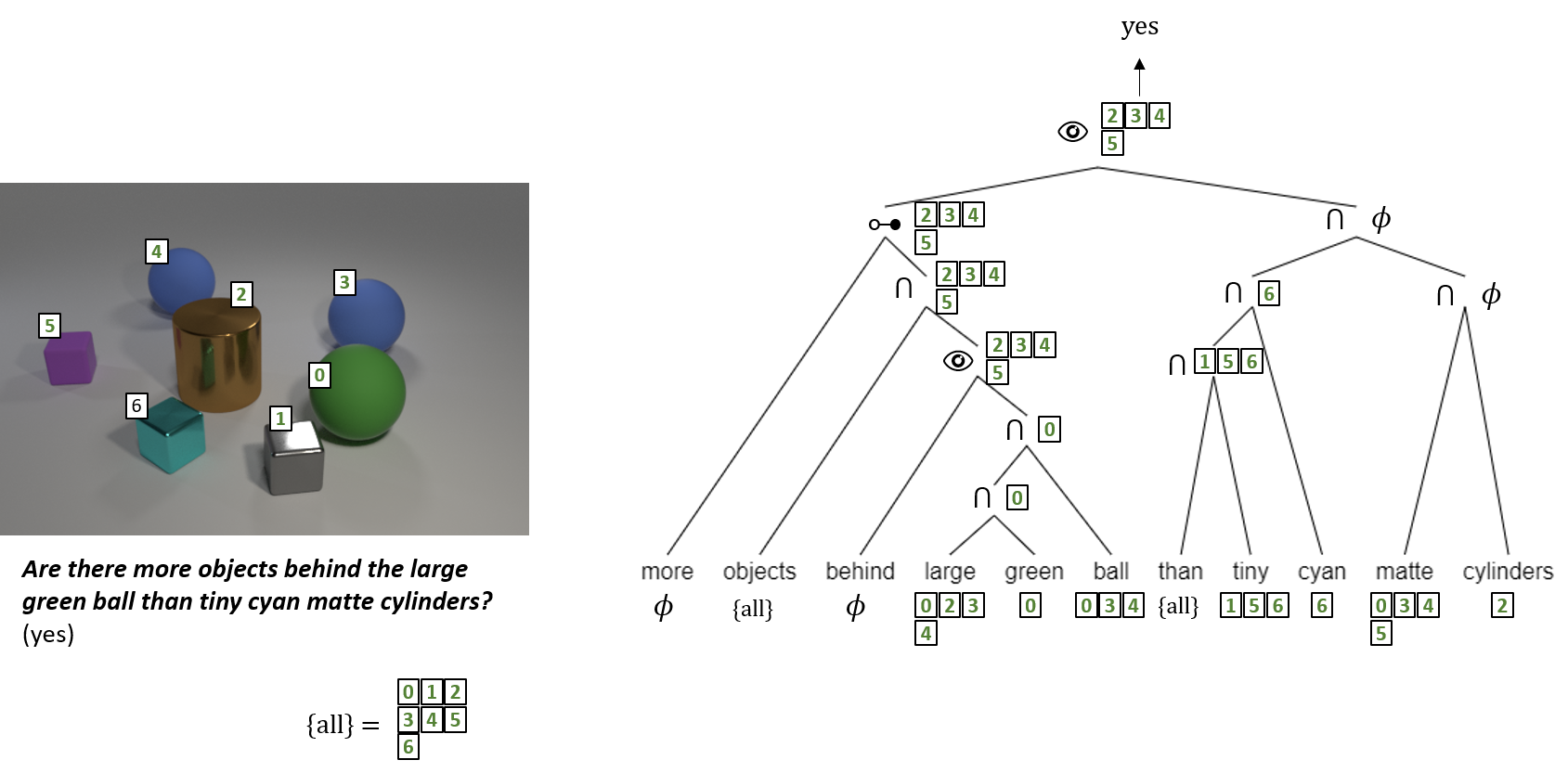}
  \caption{An example from \clevr{}.
  }
  \label{fig:ex5}
\end{figure*}

\end{document}